\documentclass{bmvc2k}

\definecolor{mydarkgreen}{RGB}{34, 139, 34}
\newcommand{\singleappendix}[1]{%
  \appendix
  \section*{#1}
  \addcontentsline{toc}{section}{#1}
  \stepcounter{section}
}
\usepackage{hyperref}
\hypersetup{
    colorlinks=true,
    linkcolor=red,
    filecolor=magenta,
    urlcolor=blue,
    citecolor=purple,
    pdftitle={Overleaf Example},
    pdfpagemode=FullScreen,
    }
    \usepackage{newfloat}
\usepackage{listings}
\usepackage{multirow}
\usepackage{adjustbox}
\usepackage{caption}
\usepackage{subcaption}
\usepackage{tikz}
\usepackage{pgfplots}
\DeclareUnicodeCharacter{2212}{−}
\usepgfplotslibrary{groupplots,dateplot}
\usetikzlibrary{patterns,shapes.arrows}
\usepackage{lipsum}  
\usepackage{wrapfig}
\usepackage[ruled,vlined]{algorithm2e}
\usepackage{colortbl}
\usepackage{algorithmic}
\usepackage{graphicx}
\usepackage{subcaption}
\usepackage{float}
\usepackage{listings}
\definecolor{codegreen}{rgb}{0,0.6,0}
\definecolor{codegray}{rgb}{0.5,0.5,0.5}
\definecolor{codepurple}{rgb}{0.58,0,0.82}
\definecolor{backcolour}{rgb}{0.95,0.95,0.92}
\definecolor{codegreen}{rgb}{0,0.6,0}
\definecolor{codegray}{rgb}{0.5,0.5,0.5}
\definecolor{codepurple}{rgb}{0.58,0,0.82}
\definecolor{backcolour}{rgb}{0.95,0.95,0.92}
\definecolor{codegreen}{rgb}{0,0.6,0}
\definecolor{codegray}{rgb}{0.5,0.5,0.5}
\definecolor{codepurple}{rgb}{0.58,0,0.82}
\definecolor{backcolour}{rgb}{0.95,0.95,0.92}

\lstdefinestyle{mystyle}{
    backgroundcolor=\color{backcolour},   
    commentstyle=\color{codegreen},
    keywordstyle=\color{magenta},
    numberstyle=\tiny\color{codegray},
    stringstyle=\color{codepurple},
    basicstyle=\ttfamily\footnotesize,
    breakatwhitespace=false,         
    breaklines=true,                 
    captionpos=b,                    
    keepspaces=true,                 
    numbers=left,                    
    numbersep=5pt,                  
    showspaces=false,                
    showstringspaces=false,
    showtabs=false,                  
    tabsize=2
}

\title{AggSS: An Aggregated
Self-Supervised Approach for Class-Incremental Learning}
\addauthor{Jayateja Kalla}{jayatejak@iisc.ac.in}{}
\addauthor{Soma Biswas}{somabiswas@iisc.ac.in}{}

\addinstitution{
Department of Electrical Engineering \\
 Indian Institute of Science \\
 Bangalore, India
}

\runninghead{Kalla, Biswas}{Aggregated Self-supervision for CIL}

\begin{document}

\maketitle

\begin{abstract}
This paper investigates the impact of self-supervised learning, specifically \textit{image rotations}, on various class-incremental learning paradigms. Here, each image with a predefined rotation is considered as a new class for training. At inference, all image rotation predictions are aggregated for the final prediction, a strategy we term {\em Aggregated Self-Supervision (AggSS)}.  We observe a shift in the deep neural network's attention towards intrinsic object features as it learns through AggSS strategy. This learning approach significantly enhances class-incremental learning by promoting robust feature learning. AggSS serves as a plug-and-play module that can be seamlessly incorporated into any class-incremental learning framework, leveraging its powerful feature learning capabilities to enhance performance across various class-incremental learning approaches. Extensive experiments conducted on standard incremental learning datasets CIFAR-100 and ImageNet-Subset demonstrate the significant role of AggSS in improving performance within these paradigms.
\end{abstract}

\section{Introduction}
\label{sec:intro}
In recent years, \textit{incremental learning}~\cite{li2017learning_algo_logit_2, rebuffi2017icarl_algo_logit_1} has gained paramount importance in the deep learning research community. In general, humans possess an inherent capacity for continuous learning, allowing them to acquire knowledge, develop new skills, and adapt to changing circumstances. On the other hand, despite their capabilities, neural networks often encounter catastrophic forgetting~\cite{goodfellow2013empirical}, which refers to their tendency to forget previously acquired knowledge when learning new information. Given challenges such as the unavailability of old data~\cite{golab2003issues_data_continous1, krempl2014open_storage, gomes2017survey_data_continoius2}, data privacy concerns~\cite{chamikara2018efficient_privacy}, and the computational expense associated with training models from scratch~\cite{zhou2023deep_deep_cil_survey}, have spurred the need for incremental learning.

In incremental learning, the process of incorporating new knowledge\footnote{it can be new classes or it can be new samples from already learned classes} into the model is commonly referred to as a \textit{task} in the literature. At each new task, the model has access to new data samples, enabling it to actively expand its knowledge. The field of incremental learning encompasses various settings, including class-incremental learning (CIL), task-incremental learning (TIL), and domain-incremental learning (DIL). The specific focus of this paper is on the more challenging setting class-incremental learning, where the task identity is not given at test time. Where as in task incremental learning, the model has access to the task identity at test-time and in domain incremental learning at every task, model will try to adapt and learn different distributions of the same classes. 

\begin{figure}[t]
    \centering
    \includegraphics[scale=0.65]{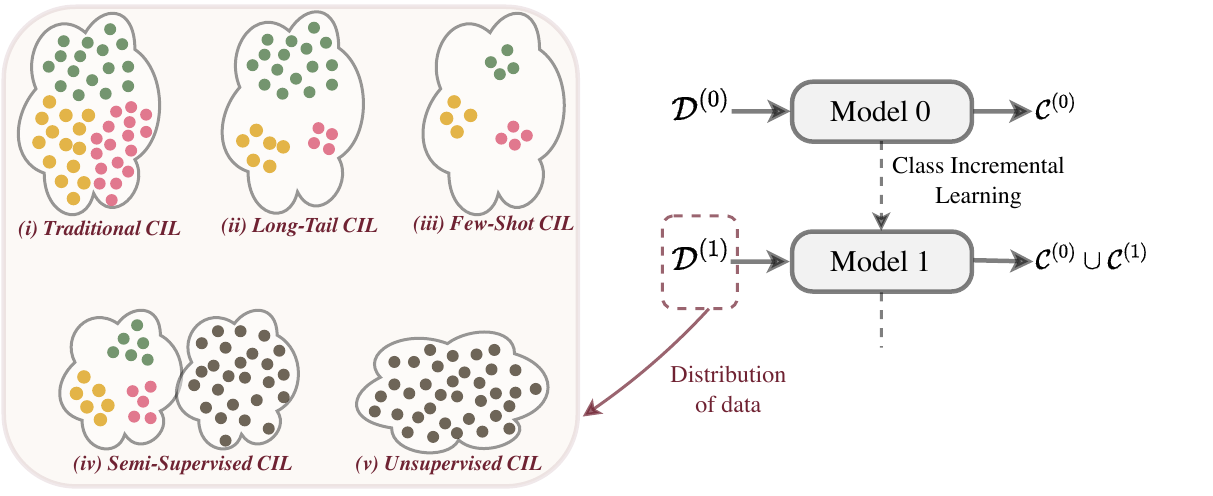}
    \vspace{0.25em}
    \caption{Illustrates various scenarios of class-incremental learning based on the available data characteristics at each incremental task: (i) Traditional CIL, where an abundant amount of labeled data is present at every task; (ii) Long-Tail CIL, where data follows long-tail distributions at every task; (iii) Few-shot CIL, where each class contains very few samples at each task; (iv) Semi-supervised CIL, where the model has access to both labeled and unlabeled data at each task; and (v) Unsupervised CIL, where every task has access only to unlabeled data.}
    \label{scenerios}
\end{figure}

In class-incremental learning~\cite{rebuffi2017icarl_algo_logit_1} initially model trained on a set of base classes, and subsequently, it is updated when a new set of classes becomes available. This allows the model to expand its knowledge and capabilities by incorporating additional classes into its existing knowledge and able to classify all the classes seen so far. CIL encompasses various scenarios depending on the characteristics of the new data available at each incremental step. These characteristics relate to how the training data is distributed across classes in each task. For example, traditional CIL~\cite{hou2019learning_algo_feat_1} involves abundant new class data at each task, few-shot CIL~\cite{tao2020few_topic_fscil2} has very few labelled samples, long-tail CIL~\cite{liu2022long} deals with a distribution of new class data with a long tail, semi-supervised CIL~\cite{kang2023soft_nncil_gsscil3} utilizes labeled and unlabeled data for each task, and unsupervised CIL~\cite{khare2021unsupervised_cil} lacks labels for all new class samples at each task. Figure \ref{scenerios} visually represents these different scenarios in CIL.

To mitigate catastrophic forgetting and integrate new information into the model, robust feature representations plays a crucial for incremental learning. Self-supervised learning (SSL)~\cite{noroozi2016unsupervised_jigsaw, gidaris2018unsupervised_rotations, chen2020simple_simclr} serves as one of the effective techniques to achieve this objective, allowing models to learn meaningful features without relying on human-annotated labels. Self-supervised label augmentation (SLA)~\cite{selfsup} have used image rotations as self-supervised and achieved good robust feature representations by treating each image and its rotation as distinct classes throughout training to enhance representation learning. For instance, a bird image rotated by $0^{\circ}$ is considered a separate class from the same bird image rotated by $90^{\circ}$. During testing, they aggregated individual classifier scores based on known image transformations to boost performance. We refer to this comprehensive training and testing process as Aggregated Self-Supervision (AggSS). A recent study by~\cite{kalla2022s3c} demonstrated the effectiveness of AggSS combined with stochastic classifiers for few-shot CIL. However, the underlying reasons for its performance improvement remain unexplored. In this paper, we delve into the qualitative and quantitative analysis of why self-supervised rotations enhance classification performance and leverage these powerful representations across diverse CIL scenarios.

To this end, this paper analyze the AggSS strategy by both qualitative and quantitative to know how its improve the feature representations. Subsequently, we aim to leverage the advantages of AggSS in various other CIL scenarios, including traditional CIL, long-tail CIL, and semi-supervised CIL. Extensive experiments on CIL datasets, such as CIFAR100~\cite{krizhevsky2009learning_cifar} and ImageNet-Subset~\cite{imagenet_cvpr09, liu2022long}, demonstrate the effectiveness of AggSS in diverse CIL scenarios.


\section{Related works}
\label{related_works_section}
In this section, we discuss related works on self-supervised learning (SSL) and various class-incremental learning (CIL) techniques. \\
\textbf{1.  Self-supervised learning (SSL)}: aims to train models on a data to extract meaningful features without relying on human-annotated labels. The term \textit{pretext} in SSL denotes that the task being solved is not the primary objective but serves as a means to generate a robust model. These approaches can be broadly categorized into three types: \textit{(i) Context-based methods:} These methods leverage the inherent relationships within the data, such as spatial structures and local/global consistency. Examples include predicting image rotations~\cite{gidaris2018unsupervised_rotations}, colorization~\cite{larsson2016learning_color1, larsson2017colorization_color2}, and solving jigsaw puzzles~\cite{noroozi2016unsupervised_jigsaw}. \textit{(ii) Contrastive learning}: This approach involves learning by contrasting similar and dissimilar pairs of examples without labels. Numerous methods have emerged in this area, including MoCo~\cite{he2020momentum_moco}, SimCLR~\cite{chen2020simple_simclr}, BYOL~\cite{grill2020bootstrap_byol}, SwAV~\cite{caron2020unsupervised_smav}, and SimSiam~\cite{chen2021exploring_simsiam}. \textit{(iii) Masked-based modeling}: With the advancements in Vision Transformers~\cite{dosovitskiy2020image_vit}, techniques like BEiT~\cite{bao2021beit_beit}, DINO~\cite{caron2021emerging_dino}, MAE~\cite{he2022masked_mae}, CAE~\cite{chen2024context_cae}, and SimMIM~\cite{xie2022simmim} have emerged that learn by predicting masked regions in images without human supervision. In our work, we analyze image rotations and their significance in class-incremental learning. \\
2. \textbf{class-incremental learning (CIL)}: aims to continuously build a comprehensive classifier that recognizes all classes encountered so far, and broadly categorized into three groups: \textit{(i) Data-centric methods}~\cite{lopez2017gradient_gem_data_centric1, chaudhry2018efficient_agem_data_centric2, isele2018selective_experience_data_centric4,castro2018end_cil1, belouadah2019il2m_cil2, ahn2021ss_cil3, wang2022foster} concentrate on solving CIL with exemplars, i.e., storing samples from old tasks, enabling the model to review former classes and resist forgetting. Recently, some exemplar free CIL approaches~\cite{zhu2021prototype_pass_exemplar_free_cil1, zhu2022self_exemplar_free_cil2,petit2023fetril_exemplar_free_cil3} have emerged where data privacy is a major constraint in CIL. \textit{(ii) Model-centric CIL methods}~\cite{yoon2017lifelong_model_centric1, yan2021dynamically_model_centric2, rusu2016progressive_model_centric_3, wang2022learning_model_centric_4, wang2022dualprompt_model_centric_5, smith2023coda_model_centric_6} mainly concentrate on model evolution in the learning process. On the other hand, parameter regularization methods~\cite{chaudhry2018riemannian_reimAN_walk_data_centric3, zenke2017continual_SI_reg_1, aljundi2018memory_reg_2, kirkpatrick2017overcoming_reg_3} estimate the importance of parameters and regularize important ones to prevent them from drifting away. \textit{(iii) Algorithm-centric CIL methods}~\cite{rebuffi2017icarl_algo_logit_1} focus on designing algorithms to maintain the model's knowledge in former tasks.  Knowledge distillation~\cite{hinton2015distilling_kd_hinton} is a popular algorithmic-centric approach that enables knowledge transfer from a teacher model to the student model. Various distillation techniques exist, such as logit distillation~\cite{rebuffi2017icarl_algo_logit_1, li2017learning_algo_logit_2, hou2018lifelong_algo_logit_3}, feature distillation~\cite{hou2019learning_algo_feat_1, dhar2019learning_algo_feat_2, kang2022class_algo_feat_3, douillard2020podnet_algo_feat_4}, and relational distillation~\cite{gao2022r_algo_relation_1, yu2020semantic_algo_relation_2} to transfer knowledge form teacher to student. \\
3. \textbf{Few-shot class-incremental learning (FS-CIL)}: learn new tasks with only a few labeled examples while retaining knowledge of previously learned ones. In FS-CIL overfitting to new tasks is a major challenge along with catastrophic forgetting due to limited number of samples. Several approaches have been proposed to address these challenges. Tao et al.~\cite{tao2020few_topic_fscil2} introduced the TOPIC neural network architecture to preserve the feature topology of both base and new classes. Other works~\cite{peng2022few_fscil3, kang2022soft_fscil4, qiu2023semantic_fscil5, kalla2022s3c} leverage various techniques to combat forgetting and overfitting. Our work draws inspiration from the S3C~\cite{kalla2022s3c} approach, which utilizes the AggSS principle and stochastic classifiers to successfully tackle issues in FS-CIL. \\
4. \textbf{Semi-supervised class-incremental learning (SS-CIL)}: raises unique challenges within CIL. Where, each task relies on partially labeled data to learn form vast amount of unlabelled data. Traditional semi-supervised~\cite{ pseudo2013simple_ss2,tarvainen2017mean_ss4, sohn2020fixmatch_ss3, chen2022debiased_ss1} approaches effectively balance labeled and unlabeled data in static settings but struggle with the continual nature of SS-CIL~\cite{kang2023soft_nncil_gsscil3}. Approaches like ORDisCo~\cite{wang2021ordisco_gsscil1}, CCIC~\cite{boschini2022continual_ccic_gsscil2}, NNSCL~\cite{kang2023soft_nncil_gsscil3} address the challenges in SS-CIL. Unlike previous methods assuming task-specific unlabeled data, ESPN~\cite{kalla2023generalized} allows unlabeled data to encompass samples from the current, previous, or even entirely unrelated tasks (outliers). In this work, we leverage ESPN as a baseline to showcase the effectiveness of our AggSS approach in the SS-CIL setting in the presence of outliers. \\
5. \textbf{Long-tail class-incremental learning (LT-CIL)}: Recently, Liu et al.~\cite{liu2022long} introduced LT-CIL using a two-stage approach and GVAlign~\cite{kalla2024robust} proposed tuning classifiers with global variance to solve issues in  LT-CIL.

\section{Aggregrated Self-Supervision}
\label{AggSS_section}
In this section, we first introduce the notations used in CIL. Next, we delve into the details of AggSS training and testing. Finally, we explore the integration of AggSS into existing CIL approaches. \\
\textbf{Notations:} In CIL, the model is initially trained on a set of base classes denoted by $\mathcal{D}^{(0)}$, which contains corresponding classes in $\mathcal{C}^{(0)}$. The incremental learning process involves subsequent updates to the model as new sets of classes become available. The data available for these incremental tasks are represented by $\{\mathcal{D}^{(1)}, \mathcal{D}^{(2)}, \dots, \mathcal{D}^{(\mathcal{T})}\}$, and their corresponding label spaces are denoted as $\{\mathcal{C}^{(1)}, \mathcal{C}^{(2)}, \dots, \mathcal{C}^{(\mathcal{T})}\}$, where $\mathcal{T}$ is the total number of incremental tasks. An important assumption in CIL is there is no overlap in the classes between different incremental tasks, meaning $\mathcal{C}^{(t)} \cap \mathcal{C}^{(s)} = \emptyset$ for $t \neq s$. Once the model has learned from the data $\mathcal{D}^{(t)}$ for each task $t$, it is expected to perform well on the classification of all the classes seen so far, which includes the union of all previously encountered classes, i.e., $\{\mathcal{C}^{(0)} \cup \mathcal{C}^{(1)} \cup \dots \cup \mathcal{C}^{(t)}\}$. In terms of model parameters, the new classifiers $\psi^{(t)}$ were added after learning task $t$ to classify the new classes.

The data characteristics at every task $\mathcal{D}^{(t)}$ make different scenarios in CIL. In traditional CIL all the classes in set $\mathcal{C}^{(t)}$ have equal and abudant number of samples, In few-shot CIL, the classes in $\mathcal{C}^{(t)}$ have few number of number of samples. In long-tail CIL, the number of samples follows the long-tail distribution. In semi-supervised CIL, data at each task $\mathcal{D}^{(t)} \in \mathcal{D}^{(t)}_{l} \cup \mathcal{D}^{(t)}_{ul}$ comprises both labeled and unlabeled data. In next section, we discuss about the AggSS training and testing details.

\subsection{AggSS training}
\begin{figure*}[t]
    \centering
     
        \includegraphics[width=\linewidth]{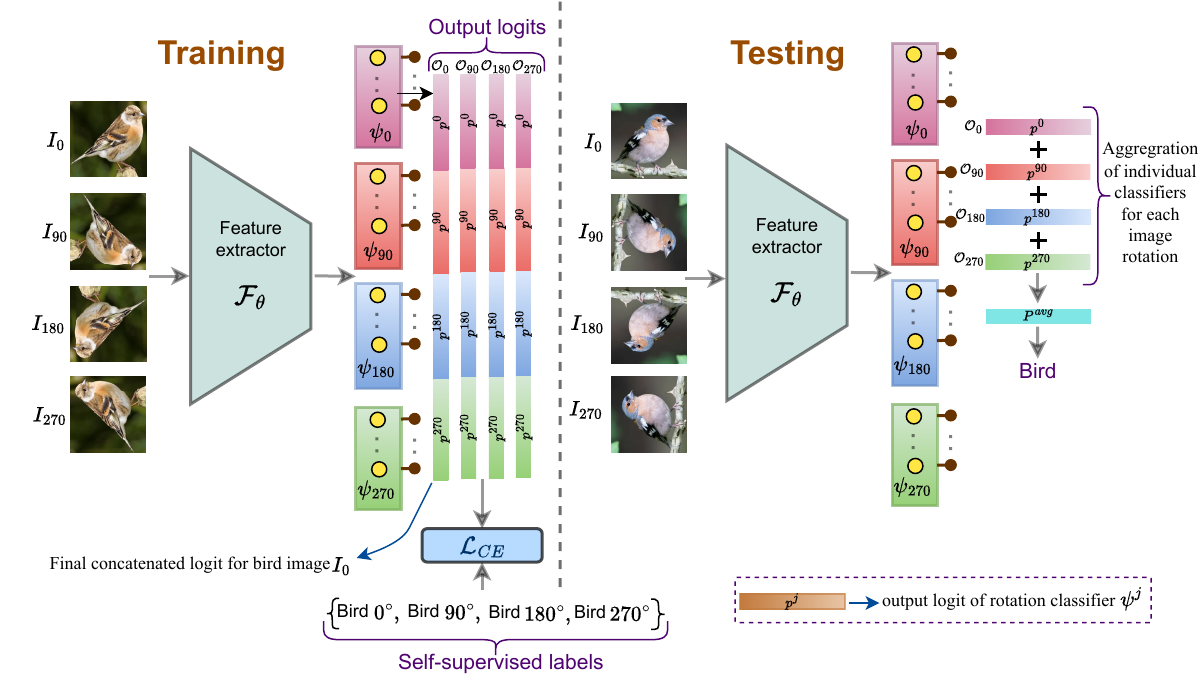}
        \caption{Illustrates both training and testing strategy of AggSS.}
        \label{Aggss_train_test_figure}

\end{figure*}
The network architecture consists of a feature extractor $\mathcal{F}_{\theta}$ and a classifier $\psi$. In the conventional cross-entropy (CE) training process, the network takes an input image $x$ along with its corresponding label $y$. The output logit is computed for input image $x$ as denoted as $p_{x} = \psi(\mathcal{F}_{\theta}(x))$. Subsequently, the CE loss is calculated for $K$ classes as $\mathcal{L}_{ce}(x,y) = - \sum_{k=1}^{K} y_{k} \log {(p_{x})_{k}}$.

During AggSS training, each transformed input image, $x_{r} = t_{r}(x)$, obtained by image transformation function $\{t_{r}\}_{r=1}^{M}$, is treated as a separate class. The corresponding label, $y$, is transformed to $y_{r} = \hat{t}_{r}(y)$ using the label transformation function $\{\hat{t}_{r}\}_{r=1}^{M}$, where $M$ represents predefined transformations (in this work, $M=4$, with images rotated by ${0^{\circ},90^{\circ},180^{\circ},270^{\circ}}$). 
Essentially in AggSS, the network learns $M$ individual transformation classifiers $\{\psi_{1},\psi_{2} ,..,\psi_{M}\}$ for each transformation. This will increase the number of classifiers in the model by $M$ times. For instance, if it originally had $K$ classifiers, now, in AggSS it will have $K \times M$ classifiers. The output logit for the input image $x_{r}$ is given by $\mathcal{O}_{x_{r}} = [p^{1}_{x_{r}}|\ p^{2}_{x_{r}}|\ ... |\ p^{M}_{x_{r}}]$, where $|$ represents concatenation, and $p^{i}_{x_{r}} = \psi_{i}(\mathcal{F}_{\theta}(x_{r}))$ is the $i^{th}$ transformation classifiers' outputs. The final AggSS training loss for a given image label pair $(x,y)$ is given by 
\begin{equation}\mathcal{L}_{AggSS}(x,y) = - \sum_{r=1}^{M} \sum_{k=1}^{K*M} (y_{r})_{k} \log {(\mathcal{O}_{x_{r}})_{k}}
\label{eq:agg_ss}
\end{equation}
\subsection{AggSS inference}
During inference, we know exactly the transformations applied to the images, we exploit the individual classifiers for effective prediction. For a given test sample $\bar{x}$, the image is augmented with all M transformations and passed through the model. Then we collect the outputs of individual rotations classifiers for respective transformations. The final aggregated logit is calculated as:

\begin{equation}
    p^{agg}_{\bar{x}} = \sum_{r=1}^{M} \psi_{r}(\mathcal{F}_{\theta}(\bar{x}_{r}))
\end{equation} 
The final prediction for the test image is $\hat{y} = argmax \ p^{agg}_{\bar{x}}$. Figure~\ref{Aggss_train_test_figure} illustrate the complete training and testing process of the AggSS. 


\subsubsection{Incorporating AggSS into CIL approaches}
The AggSS approach can function as a plug-and-play module for any image classification training framework, although in this study, we restrict its application to incremental learning approaches. In all incremental learning scenarios, there exists a cross-entropy loss coupled with augmented losses aimed to address challenges in CIL:
\begin{equation}
    \mathcal{L}_{inc} = \mathcal{L}_{ce} + \mathcal{L}_{aug}
\end{equation}

Approaches such as UCIR~\cite{hou2018lifelong_algo_logit_3} incorporate a augmented lossees comprising feature distillation and margin ranking loss, while in PODNET~\cite{douillard2020podnet_algo_feat_4}, the augmented loss encompasses pooled distillation loss. AggSS can be directly integrated into the CIL paradigm by replacing the cross-entropy loss. This integration primarily impacts the model architecture, increasing the number of classifiers based on the chosen transformations. Consequently, the loss at each incremental task becomes:
\begin{equation}
    \mathcal{L}_{inc} = \mathcal{L}_{AggSS} + \mathcal{L}_{aug}
\end{equation}

\section{Experiments}
\label{Expts}
In this section, we discuss the dataset and implementation details, and experimental results of AggSS on various CIL settings.
\subsection{Datasets and implementation details:} 
To evaluate the effectiveness of AggSS, we conducted experiments on two widely used benchmark datasets in class-incremental learning. \textbf{(i) CIFAR-100}: This dataset comprises 50,000 training images and 10,000 test images, each with dimensions of 32x32 pixels. It encompasses 100 classes. \textbf{(ii) ImageNet-Subset}: This subset includes 100 classes sampled from the larger ImageNet dataset~\cite{imagenet_cvpr09}. All images were resized to 256x256 pixels and randomly cropped to 224x224 pixels during the training phase. 

We used the same task splits as LWS~\cite{liu2022long} for both traditional and long-tail CIL settings. In the 5-task configuration $(\mathcal{T}=5)$, we progressively introduced 10 new classes during each incremental task (i.e., $(50-10-\ldots-10)$). Similarly, in the 10-task setup $(\mathcal{T}=10)$, we incorporated 5 new classes in each incremental task (i.e., $(50-5-\ldots-5)$). For semi-supervised class-incremental learning, we followed the splits proposed in ESPN~\cite{kalla2023generalized}. For the 4-task configuration, it is similar to traditional CIL split (i.e., $50-10-10-10-10$), and for the 3-task configuration, the splits are balanced (i.e., $20-20-20-20$). The last task classes in both the scenarios were used as outliers. \\
\textbf{Model details:} We employed ResNet-32 for the CIFAR-100 dataset and ResNet-18 for the ImageNet-Subset in all approaches to ensure fair comparison. \\
\textbf{Training details:} During training, the initial learning rate was set to 0.1. For CIFAR-100, after 100 and 200 epochs (300 epochs in total), the learning rate was reduced by factor 10. For the ImageNet Subset, the learning rate began at 0.1 and was reduced by factor 10 after 30 and 60 epochs (90 epochs in total). The batch size for the CIFAR-100 experiments was set to 128, while for the ImageNet-Subset experiments, 32 was used. All the experiments are conducted on two NVIDIA RTX A5000 GPUs, each equipped with 24GB of memory, using the PyTorch deep learning library. \\
\textbf{Evaluation metrics:} We employ the widely recognized CIL evaluation metric, average incremental accuracy~\cite{rebuffi2017icarl_algo_logit_1, lopez2017gradient_gem_data_centric1} to report the results. Here, let $t$ represent the task ID, where $t\in{0,1,...,\mathcal{T}}$. We define $Acc_{0:n}^{t}$ as the model's accuracy on the test data of all tasks from $0$ to $n$ after learning task $t$, where $n \leq t$. Consequently, upon completion of task $T$, the average incremental accuracy is computed as $\frac{1}{T}\sum_{t=0}^{T} Acc_{0:t}^{t}$.
\begin{table}[t]
\centering
\begin{adjustbox}{max width=0.8\linewidth}
\begin{tabular}{lllll}
\hline \hline
\multirow{2}{*}{\textit{Method}} & \multicolumn{2}{c}{\rule{-2pt}{10pt} CIFAR 100}          & \multicolumn{2}{c}{ImageNet-Subset}    \\ \cline{2-5} 
                                 & \multicolumn{1}{l}{\rule{-2pt}{10pt} 5 tasks} & 10 tasks & \multicolumn{1}{l}{5 tasks} & 10 tasks \\ \hline \hline \rule{-2pt}{10pt} 
               UCIR~\cite{hou2019learning_algo_feat_1} (CVPR 2019)     & \multicolumn{1}{l}{61.15}        &    58.74      & \multicolumn{1}{l}{69.11}        &65.15          \\ 
   

 
 \rowcolor[HTML]{F5F5F5} UCIR + AggSS (ours)                   & \multicolumn{1}{l}{\textbf{68.75} \color[HTML]{a11f1f} 7.60 $\uparrow$  }      &       \textbf{67.21} \color[HTML]{a11f1f} 8.47 $\uparrow$  & \multicolumn{1}{l}{\textbf{74.51} \color[HTML]{a11f1f} 5.40 $\uparrow$}      &    \textbf{70.54}   \color[HTML]{a11f1f} 5.39 $\uparrow$

\\ \hline \rule{-2pt}{10pt}
 
  PODNET~\cite{douillard2020podnet_algo_feat_4} (ECCV 2020)     & \multicolumn{1}{l}{63.15}        &       61.16   & \multicolumn{1}{l}{67.92}        &    62.39      \\ 


 
 \rowcolor[HTML]{F5F5F5}PODNET + AggSS (ours)                   & \multicolumn{1}{l}{\textbf{67.16} \color[HTML]{a11f1f} 4.01 $\uparrow$}     &      \textbf{67.12} \color[HTML]{a11f1f} 5.96 $\uparrow$ & \multicolumn{1}{l}{\textbf{73.83} \color[HTML]{a11f1f} 5.91 $\uparrow$ }    &   \textbf{71.10}  \color[HTML]{a11f1f} 8.71 $\uparrow$ 
\\ \hline \rule{-2pt}{10pt}
 
  FOSTER~\cite{wang2022foster} (ECCV 2022)     & \multicolumn{1}{l}{67.67}        &       64.20   & \multicolumn{1}{l}{77.71}        &    75.60      \\ 
 
 \rowcolor[HTML]{F5F5F5} FOSTER + AggSS (ours)                   & \multicolumn{1}{l}{\textbf{70.59} \color[HTML]{a11f1f} 2.92 $\uparrow$}     &      \textbf{69.24} \color[HTML]{a11f1f} 5.04 $\uparrow$ & \multicolumn{1}{l}{\textbf{79.42} \color[HTML]{a11f1f} 1.71 $\uparrow$ }    &   \textbf{77.78}  \color[HTML]{a11f1f} 2.18 $\uparrow$ 

\\ \hline \rule{-2pt}{10pt}
   FeTriL$^{\dag}$~\cite{petit2023fetril_exemplar_free_cil3} (WACV 2023)     & \multicolumn{1}{l}{65.87}        &       64.83   & \multicolumn{1}{l}{73.85}        &    72.89      \\ 
 
 \rowcolor[HTML]{F5F5F5} FeTriL$^{\dag}$ + AggSS (ours)                   & \multicolumn{1}{l}{\textbf{69.10} \color[HTML]{a11f1f} 3.14 $\uparrow$}     &      \textbf{67.35} \color[HTML]{a11f1f} 2.52 $\uparrow$ & \multicolumn{1}{l}{\textbf{78.45} \color[HTML]{a11f1f} 4.60 $\uparrow$ }    &   \textbf{77.14}  \color[HTML]{a11f1f} 4.25 $\uparrow$ 
\\

\hline \hline
\end{tabular}
\end{adjustbox}
\vspace{0.75em}
\caption{Experimental results on traditional class-incremental learning. $\dag$ represents the exemplar-free approach, where there is no storage to save old classes data.}
\label{traditional_cil}
\end{table}

\subsection{Experiment results:}
\textbf{Traditional CIL:} Table~\ref{traditional_cil} presents the experimental results on the traditional CIL protocol. By incorporating AggSS as plug and play module over various traditional and exemplar free CIL approaches, its robust feature representations significantly enhance the performance of CIL approaches. On CIFAR-100, for both 5 tasks and 10 tasks, AggSS demonstrates a $7.60\%$ and $8.47\%$ relative improvement when UCIR is used as the baseline. Similarly, when PODNET serves as the baseline on CIFAR, AggSS shows a $4.01\%$ and $5.96\%$ improvement for 5 tasks and 10 tasks, respectively. AggSS also proves effective on the ImageNet-Subset dataset, showcasing a $5.40\%$ improvement in the 5-task scenario and a $5.39\%$ improvement in the 10-task scenario when UCIR is considered as the baseline. Moreover, it consistently outperforms FeTriL~\cite{petit2023fetril_exemplar_free_cil3} in the exemplar-free scenario. \\
\textbf{Long-tail CIL:} To assess the effectiveness of AggSS, we conducted experiments on long-tail CIL with two different scenarios (ordered and shuffled). The AggSS plug and play module demonstrated improvements in both distributions and improved over SOTA approaches. In the 5-task ordered LT scenario, it showed a $1.82\%$ improvement and a $0.76\%$ improvement in the 10-task scenario with UCIR as the baseline. On ImageNet, with UCIR as the baseline, it exhibited a $6.98\%$ improvement in the 5-task setting and a $6.39\%$ improvement in the 10-task setting compared to the state-of-the-art approach in ordered LT. The complete results are shown in Table~\ref{long_tail_cil_table} for various tasks with all datasets under different baselines.\\
\textbf{Semi-supervised CIL:} Table~\ref{semi_supervised_expt_table} displays the experimental results in semi-supervised CIL. In the 4-task scenario (50-10-10-10-10), AggSS incorporated with ESPN improves by $6.3\%$. In the 3-task scenario, which is particularly challenging due to very few base classes (20-20-20-20) making it hard to generalize, AggSS shows a $7.25\%$ relative improvement over ESPN.
In next section we discuss both qualitative and quantitative analysis of AggSS on CIFAR10 datset. In the next section, we discuss both the qualitative and quantitative analyses of AggSS, highlighting a shift in the deep neural network’s attention towards intrinsic object features as it learns through the AggSS strategy.


\begin{table*}[t]
\centering
\begin{adjustbox}{max width=\linewidth}
\begin{tabular}{ccccccccc}
\hline \hline
\textit{long tail distribution type} $\rightarrow$&\multicolumn{4}{|c}{\rule{-2pt}{10pt} Ordered long tail}                                                                                                                         & \multicolumn{4}{|c}{Shuffled long tail}                                                                         \\ \hline
\multicolumn{1}{c}{\multirow{2}{*}{\textit{Method} $\downarrow$}} & \multicolumn{2}{|c}{\rule{-2pt}{10pt} CIFAR-100}                               & \multicolumn{2}{c}{ImageNet-Subset}    & \multicolumn{2}{|c}{CIFAR-100}                               & \multicolumn{2}{c}{ImageNet-Subset}    \\ \cline{2-9} 
\multicolumn{1}{c}{}                                 & \multicolumn{1}{|l}{5 tasks} & \multicolumn{1}{l}{10 tasks} & \multicolumn{1}{l}{5 tasks} & \multicolumn{1}{l}{10 tasks} & \multicolumn{1}{|l}{\rule{-2pt}{10pt} 5 tasks} & \multicolumn{1}{l}{10 tasks} & \multicolumn{1}{l}{5 tasks} & \multicolumn{1}{l}{10 tasks} \\ \hline \hline 

\multicolumn{1}{c|}{\rule{-2pt}{10pt}  UCIR~\cite{hou2019learning_algo_feat_1} (CVPR 2019) }                                & \multicolumn{1}{l}{42.69}        & \multicolumn{1}{l}{42.15}         & \multicolumn{1}{l}{56.45}        &    \multicolumn{1}{l}{55.44}      & \multicolumn{1}{|l}{35.09}       & \multicolumn{1}{l}{34.59}         & \multicolumn{1}{l}{46.45}        & \multicolumn{1}{l}{45.31}         \\ 

\multicolumn{1}{c|}{UCIR + LWS~\cite{liu2022long} (ECCV 2022) }                                & \multicolumn{1}{l}{45.88}        & \multicolumn{1}{l}{45.73}         & \multicolumn{1}{l}{57.22}        &     \multicolumn{1}{l}{55.41}     & \multicolumn{1}{|l}{39.40}       & \multicolumn{1}{l}{39.00}         & \multicolumn{1}{l}{49.42}        &\multicolumn{1}{l}{47.96}          \\ 

\multicolumn{1}{c|}{UCIR + GVAlign~\cite{kalla2024robust} (WACV 2024)}                                & \multicolumn{1}{l}{47.13 }        & \multicolumn{1}{l}{46.82}         & \multicolumn{1}{l}{58.08}        &     \multicolumn{1}{l}{56.68}   & \multicolumn{1}{|l}{42.80}       & \multicolumn{1}{l}{41.64}         & \multicolumn{1}{l}{50.69}        &  \multicolumn{1}{l}{47.58}      \\  

\rowcolor[HTML]{F5F5F5} \multicolumn{1}{c|}{UCIR + AggSS (ours)}                                & \multicolumn{1}{l}{\textbf{48.95} \color[HTML]{a11f1f}  1.82 $\uparrow$ }    & \multicolumn{1}{l}{\textbf{47.58} \color[HTML]{a11f1f}  0.76 $\uparrow$}       & \multicolumn{1}{l}{\textbf{65.06} \color[HTML]{a11f1f}  6.98 $\uparrow$}     &    \multicolumn{1}{l}{\textbf{63.07} \color[HTML]{a11f1f}  6.39 $\uparrow$}    & \multicolumn{1}{|c}{\textbf{43.18} \color[HTML]{a11f1f}  0.38 $\uparrow$}      & \multicolumn{1}{c}{\textbf{42.86} \color[HTML]{a11f1f}  1.22 $\uparrow$}         & \multicolumn{1}{c}{\textbf{56.06} \color[HTML]{a11f1f} 5.37 $\uparrow$  }     &     \textbf{55.71  } \color[HTML]{a11f1f}  7.75 $\uparrow$ \\ 
\hline 

\multicolumn{1}{c|}{\rule{-2pt}{10pt}  PODNET~\cite{douillard2020podnet_algo_feat_4} (ECCV 2020) }                                & \multicolumn{1}{l}{44.07}        & \multicolumn{1}{l}{43.96}         & \multicolumn{1}{l}{59.16}        &   \multicolumn{1}{l}{57.74}       &  \multicolumn{1}{|l}{36.64}       & \multicolumn{1}{l}{34.84}         & \multicolumn{1}{l}{47.61}        & \multicolumn{1}{l}{47.85}         \\ 

\multicolumn{1}{c|}{PODNET + LWS~\cite{liu2022long} (ECCV 2022) }                                & \multicolumn{1}{l}{44.38}        & \multicolumn{1}{l}{44.35}         & \multicolumn{1}{l}{60.12}        & \multicolumn{1}{l}{59.09}         & \multicolumn{1}{|l}{36.37}       & \multicolumn{1}{l}{37.03}         & \multicolumn{1}{l}{49.75}        &\multicolumn{1}{l}{49.51}          \\ 

\multicolumn{1}{c|}{PODNET + GVAlign~\cite{kalla2024robust} (WACV 2024)}                                & \multicolumn{1}{l}{48.41}        & \multicolumn{1}{l}{47.71}         & \multicolumn{1}{l}{61.06}        &   \multicolumn{1}{l}{60.08}       & \multicolumn{1}{|l}{42.72}       & \multicolumn{1}{l}{41.61}         & \multicolumn{1}{l}{52.01}        &    \multicolumn{1}{l}{50.81} \\ 

\rowcolor[HTML]{F5F5F5} \multicolumn{1}{c|}{PODNET + AggSS (ours)  }                                & \multicolumn{1}{l}{\textbf{52.93} \color[HTML]{a11f1f} 4.52 $\uparrow$ }        & \multicolumn{1}{l}{\textbf{51.97} \color[HTML]{a11f1f} 4.26 $\uparrow$ }         & \multicolumn{1}{l}{\textbf{66.38} \color[HTML]{a11f1f} 5.32 $\uparrow$ }        &    \multicolumn{1}{l}{\textbf{65.01} \color[HTML]{a11f1f} 4.93 $\uparrow$ }      & \multicolumn{1}{|l}{\textbf{43.80} \color[HTML]{a11f1f}  1.08 $\uparrow$ }       & \multicolumn{1}{l}{\textbf{44.78} \color[HTML]{a11f1f} 3.17 $\uparrow$ }         & \multicolumn{1}{l}{\textbf{58.38} \color[HTML]{a11f1f} 6.37 $\uparrow$ }        &    \multicolumn{1}{l}{\textbf{58.14} \color[HTML]{a11f1f} 7.33 $\uparrow$ }      \\ 

\hline 
\hline 

\end{tabular}
\end{adjustbox}
\vspace{0.75em}
\caption{Experimental results on long tail class-incremental learning ($\uparrow$ indicates the relative improvement from second best results).}
\label{long_tail_cil_table}
\end{table*}
\begin{figure}[t]
  \centering
  \begin{minipage}[b]{0.45\textwidth}
  \centering
  \begin{adjustbox}{max width=1\linewidth}
  \begin{tabular}{lll}
\hline \hline
\multirow{2}{*}{\textit{Method}} & \multicolumn{2}{c}{\rule{-2pt}{10pt} CIFAR 100}             \\ \cline{2-3} 
                                 & \multicolumn{1}{l}{\rule{-2pt}{10pt} 4 tasks} & 3 tasks  \\ \hline \hline \rule{-2pt}{10pt} 
   
               UCIR~\cite{hou2019learning_algo_feat_1} (CVPR 2019)     & \multicolumn{1}{l}{58.99}        &    \multicolumn{1}{l}{60.52}            \\ 
               PODNET~\cite{douillard2020podnet_algo_feat_4} (ECCV 2020)     & \multicolumn{1}{l}{63.38}        &    \multicolumn{1}{l}{58.03}           \\ 
               ESPN~\cite{kalla2023generalized} (MTA 2023)     & \multicolumn{1}{l}{64.01}        &    \multicolumn{1}{l}{64.11}           \\ 
              \rowcolor[HTML]{F5F5F5} ESPN + AggSS (Ours)     & \multicolumn{1}{l}{\textbf{70.32}  \color[HTML]{a11f1f} 6.30 $\uparrow$}        & \multicolumn{1}{l}{\textbf{71.36}  \color[HTML]{a11f1f} 7.25  $\uparrow$}          \\

\hline \hline
\end{tabular}
\end{adjustbox}
\vspace{1em}
  \caption{Experiment results on semi-supervised CIL setting.}
  \label{semi_supervised_expt_table}
  \end{minipage}
  \hfill
  \begin{minipage}[b]{0.53\textwidth}
    \centering
    \begin{tikzpicture}[scale=1]
\definecolor{color0}{HTML}{F0F8FF}
\definecolor{color1}{HTML}{778899}
\definecolor{color2}{HTML}{696969}
  \begin{axis}[
    xlabel={Image Rotations},
    ylabel={Accuracy (\%)},
    xtick={1,2,3,4},
    xticklabels={1,2,4,8},
    ymin=93, ymax=96,
    legend style={at={(0.5,-0.15)}, anchor=north, legend columns=-1},
    axis background/.style={fill=color0},
    width=5cm, height=3cm,
   x label style={at={(axis description cs:0.5,-0.0)},anchor=north},
    y label style={at={(axis description cs:0.05,0.5)},anchor=north},
    ]
    
    \addplot[very thick, color= color1, mark=*,
      mark options={scale=2, color=color2},
      ] coordinates {
        (1, 93.3)
        (2, 94.7)
        (3, 95.8)
        (4, 95.6)
      };
    
  \end{axis}
\end{tikzpicture}
    \vskip 0.1in
    
    \caption{Performance vs Rotations}
\label{rotations_plot}
  \end{minipage}
\end{figure}

\section{Analysis of AggSS}
\begin{figure}
    \centering
    \includegraphics[width=\linewidth]{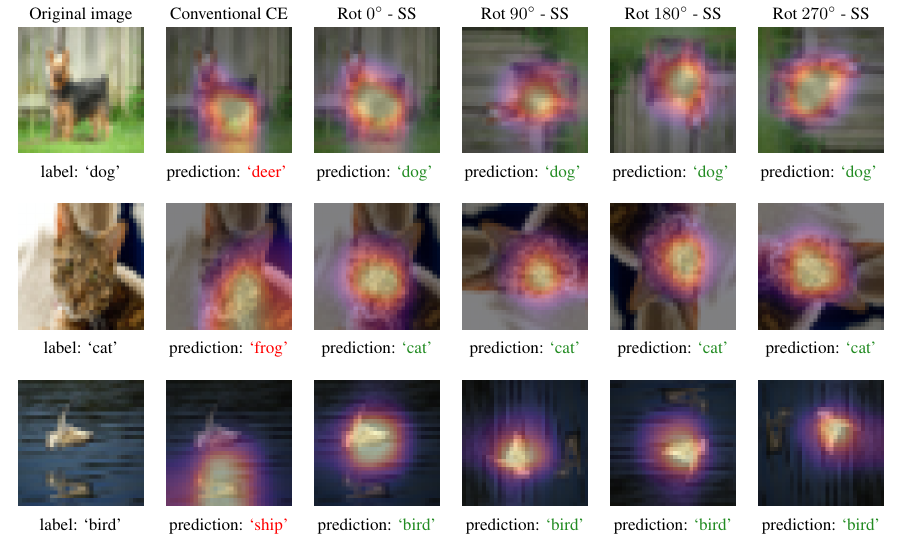}
    \vspace{0cm}
    \caption{Displays GradCAM patterns for images trained using both conventional cross-entropy (CE) loss and AggSS loss. In AggSS, each rotation is treated as a separate class.}
    \label{Gradcam_maps}
    
\end{figure}

To comprehensively analyze the performance of AggSS both qualitatively and quantitatively, we conducted an analysis experiment using the CIFAR10 dataset. In this experiment, we trained two ResNet18 models: one using the traditional CE loss and the other using the AggSS training procedure.\\
\textbf{Quantitative Analysis:}
The model trained with conventional CE achieved an accuracy of $93.30\%$, while the AggSS-trained model outperformed it significantly with an accuracy of $95.81\%$. This substantial difference of $2.51\%$ unequivocally demonstrates the clear advantage of AggSS over traditional training quantitatively.  \\
\textbf{Qualitative Analysis:}
We conducted a more qualitative analysis to gain deeper insights into the performance disparities between the two models. For this purpose, we employed GradCAM~\cite{selvaraju2017grad_cam} to analyze the network's attention patterns and to identify the regions in the images influencing the model's decision-making process. The findings from the GradCAM analysis are as follows:
\begin{itemize}
    \item 
In the first row of figure~\ref{Gradcam_maps}, when presented with an original image of a dog, the conventional CE model focuses predominantly on the dog's legs region and incorrectly predicts it as a deer. On the contrary, the AggSS-trained model, utilizing different rotations of the image to learn separate classifiers, focuses on distinct parts of the dog, ultimately resulting in an aggregated prediction of a dog.
\item
In the second row, when given an image of a cat, the conventional CE model attends to the cat's checks and mistakenly predicts it as a frog. In contrast, the AggSS rotations training procedure directs the model's attention to the actual face of the cat, leading to an accurate prediction of it being a cat.
\item 
In the third row, the presence of a bird's shadow in water causes the conventional CE model to misclassify it as a ship. However, AggSS directs its attention to the bird itself, correctly predicting its class as bird.
\end{itemize}

The GradCAM analysis and the quantitative findings firmly support the conclusion that AggSS outperforms traditional CE training in both qualitative and quantitative aspects. A pertinent question arises regarding the optimal number of rotations to consider. As depicted in Figure~\ref{rotations_plot}, the model's performance remains stable until four rotations, after which a slight dip in performance is observed.

\section{Conclusion}
In conclusion, this paper has examined the impact of self-supervised learning, specifically image rotations, on various CIL paradigms. Through the novel of Aggregated Self-Supervision (AggSS), where each image with a predefined rotation is treated as a new class for training and aggregated for final prediction, we observed a notable shift in the deep neural network's attention towards intrinsic object features. This strategy significantly enhances feature learning representations of the deep neural models. AggSS serves as a versatile plug-and-play module that can also seamlessly integrate into any CIL framework, leveraging its powerful feature learning capabilities to enhance performance across diverse CIL tasks. Our extensive experiments conducted on standard incremental learning datasets CIFAR100 and ImageNet-Subset provide compelling evidence of the significant role played by AggSS in improving performance within these paradigms. \\
\textbf{Limitations and Future Directions:}
A primary limitation of AggSS lies in the increase in the number of classifiers with a corresponding increase in transformations. A promising avenue for future research involves exploring strategies to achieve comparable attention features with a limited set of transformations, thereby mitigating the rise in the number of classifiers. Furthermore, investigating the behavior of vision transformer architectures presents an intriguing area for future exploration.

\singleappendix{Appendix}
\subsection{Algorithm}
Algorithm~\ref{Algorithm_aggss_in_cil} outlines the procedure for integrating AggSS into CIL approaches. Initially, the base task is trained with the AggSS loss instead of the cross-entropy (CE) loss. During incremental steps, model parameters are updated using the AggSS loss along with augmented losses proposed by CIL approaches to address the challenges inherent in CIL.
\begin{algorithm}[h]
\small
    \caption{Incorporating AggSS into CIL approaches}
    \KwIn{$\{ \mathcal{F}_{\theta}, \psi^{(0)}_{1:M}\}\leftarrow$ Initial model \\  
    $\{ \mathcal{D}^{(0)}, \mathcal{D}^{(1)},..,\mathcal{D}^{(T)} \} \leftarrow $ Data stream\\
    $e$$\leftarrow$ No.of epochs \\ 
    $\mathcal{E} = \{ \}$ $\leftarrow$ Empty exemplar buffer\;}
   
    \For{$t \leftarrow 0$ to $T$}{
    
        $\mathcal{D}^{(t)} = \{ x_{i}, y_{i}  \}_{i=1}^{N_{t}}$ \\
        
        \For{$\text{epoch} \gets 1$ to $e$}{
        
        $\mathcal{B} = $ SampleMiniBatch$(\mathcal{D}^{(t)} \cup \mathcal{E})$ \\
        $\mathcal{\hat{B}} \leftarrow \text{ImageTransfomations}(\mathcal{B}) $\\
        
        \If{t=0}{
        $\mathcal{O}^{(0)} = \psi^{(0)}_{1:M}(\mathcal{F}_{\theta}(\mathcal{\hat{B}})) \text{{\scriptsize{\tcp{Image passed through all rotation classifiers}}}}$ \\
         $\mathcal{L}_{AggSS}=\text{AggSSLoss}(\mathcal{\hat{B}}, \mathcal{O}^{(0)}) \text{{\scriptsize{\tcp{Aggregated SS loss as in Eq. ~\ref{eq:agg_ss}}}}} $\\
         $\{ \mathcal{F}_{\theta}, \psi^{(0)}_{1:M}\} \leftarrow$ UpdateParameters$(\mathcal{L}_{AggSS})$

        }
       
        \If{t$>$0}{
        $\mathcal{O}^{(0:t)} = \psi^{(0:t)}_{1:M}(\mathcal{F}_{\theta}(\mathcal{\hat{B}})) \text{{\scriptsize{\tcp{Image passed through all rotation classifiers}}} }$\\
        $\mathcal{L}_{AggSS} = $ AggSSLoss$(\mathcal{\hat{B}}, \mathcal{O}^{(0:t)}) \text{{\scriptsize{\tcp{Aggregated SS loss as in Eq. ~\ref{eq:agg_ss}}}}} $ \\
         $\mathcal{L}_{aug} = \text{AugmentedLosses}(\mathcal{\hat{B}}, \mathcal{O}^{(0:t)}) \text{{\scriptsize{\tcp{Additional augmented losses }}}} $ \\
         $\{ \mathcal{F}_{\theta}, \psi^{(0:t)}_{1:M}\} \leftarrow$ UpdateParameters$(\mathcal{L}_{AggSS} + \mathcal{L}_{aug})$
        }
        
        }

     $\mathcal{E} \leftarrow \text{UpdateExemplars}(\mathcal{D}^{(t)})$
   } 
    
    \Return{$\{ \mathcal{F}_{\theta}, \psi^{(0:t)}_{1:M}\}$}
    \label{Algorithm_aggss_in_cil}
\end{algorithm}
\subsection{PyTorch Sample Code for Incorporating AggSS into Model Training and Inference for CIL Algorithms}

Sample PyTorch training code for incorporating AggSS into model training is provided in Listing~\ref{lst:pytorch_code_train}, while the corresponding inference code is given in Listing~\ref{lst:pytorch_code_test}.
\label{pytorch_train_test_appendix}
\begin{lstlisting}[language=Python, caption={Incorporating AggSS in training}, label={lst:pytorch_code_train},  basicstyle=\scriptsize]
############ Training #############
for batch_idx, (inputs, targets) in enumerate(trainloader):
        inputs, targets = inputs.to(device), targets.to(device)
        H, W = inputs.shape[-1], inputs.shape[-2]
        inputs = torch.stack([torch.rot90(inputs, k, (2, 3)) for k in range(4)], 1)
        inputs = inputs.view(-1, 3, H, W)
        targets = torch.stack([targets * 4 + k for k in range(4)], 1).view(-1)
        
        optimizer.zero_grad()
        outputs = net(inputs)
        loss = criterion(outputs, targets)
        loss.backward()
        optimizer.step()
\end{lstlisting}

\begin{lstlisting}[language=Python, caption={Incorporating AggSS at inference}, label={lst:pytorch_code_test},  basicstyle=\scriptsize]
############ Inference #############
for batch_idx, (inputs, _) in enumerate(testloader):
        inputs = inputs.to(device)
        H, W = inputs.shape[-1], inputs.shape[-2]
        inputs = torch.stack([torch.rot90(inputs, k, (2, 3)) for k in range(4)], 1)
        inputs = inputs.view(-1, 3, H, W)
        outputs = net(inputs)
        AG = 0.
        for k in range(4):
            AG = AG + outputs[k::4, k::4] / 4.
        _, predicted = AG.max(1)
\end{lstlisting}

\subsection{Illustration of data distributions:}
\begin{figure*}[h]
\centering
\scalebox{.73}{\input{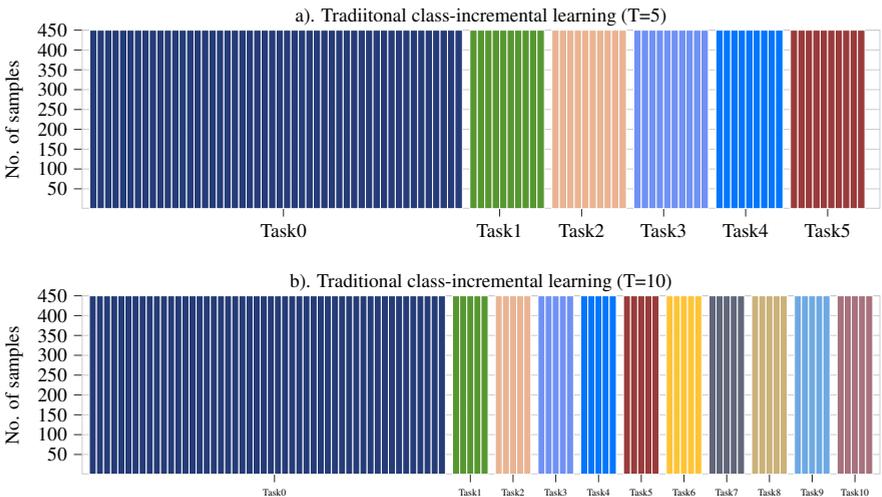}}
\vspace{1em}
\caption{CIFAR100 data distributions class-incremental learning for T=5 and T=10 tasks.}

\label{cil_data_distributions_appendix}
\end{figure*}

\begin{figure}[H]
\centering
\scalebox{.75}{\input{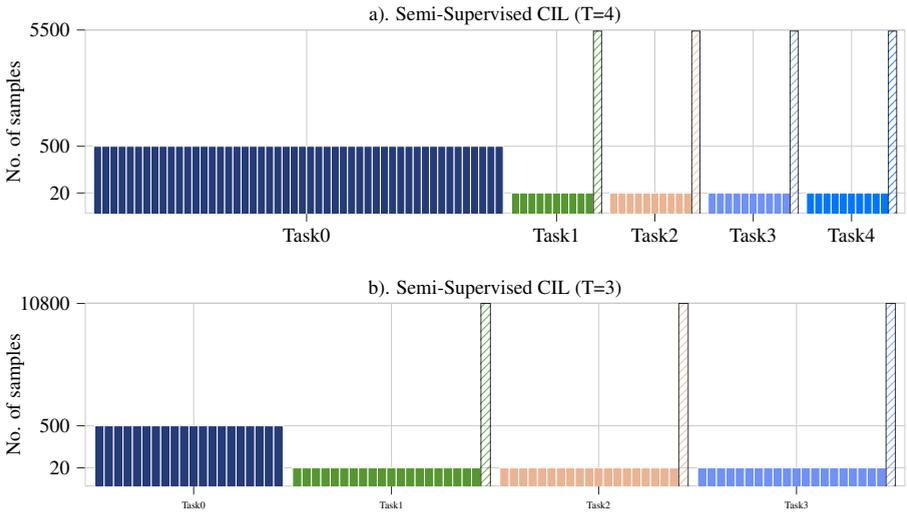}}
\vspace{1em}
\caption{CIFAR100 data distributions semi-supervised CIL for T=4 and T=3 tasks. The slanted lines inside bars indicate the presence of unlabeled data available at each task.}

\label{sscil_data_distributions_appendix}
\end{figure}

In this section, we present visualizations of the class-wise distribution of CIFAR100 datasets across various Class-Incremental Learning (CIL) paradigms. Figure~\ref{cil_data_distributions_appendix} displays the traditional CIL scenario, while Figure~\ref{ltcil_shuffled_data_distributions_appendix} depicts the shuffled long-tail CIL distribution. Additionally, Figure~\ref{ltcil_ordered_data_distributions_appendix} showcases the ordered long-tail CIL distribution, and Figure~\ref{sscil_data_distributions_appendix} illustrates the data distribution settings for Semi-Supervised CIL.

\begin{figure*}[h]
\centering
\scalebox{.73}{\input{shuffled_long_tail_data}}
\vspace{1em}
\caption{CIFAR100 data distributions long-tail class-incremental learning (shuffled) for T=5 and T=10 tasks.}
\label{ltcil_shuffled_data_distributions_appendix}

\end{figure*}

\begin{figure}[H]
\centering
\scalebox{.75}{\input{ordered_long_tail_data}}
\vspace{1em}
\caption{CIFAR100 data distributions long-tail class-incremental learning (ordered) for T=5 and T=10 tasks..}
\label{ltcil_ordered_data_distributions_appendix}

\end{figure}

\bibliography{egbib}

\begin{thebibliography}{71}
\providecommand{\natexlab}[1]{#1}
\providecommand{\url}[1]{\texttt{#1}}
\expandafter\ifx\csname urlstyle\endcsname\relax
  \providecommand{\doi}[1]{doi: #1}\else
  \providecommand{\doi}{doi: \begingroup \urlstyle{rm}\Url}\fi

\bibitem[Ahn et~al.(2021)Ahn, Kwak, Lim, Bang, Kim, and Moon]{ahn2021ss_cil3}
Hongjoon Ahn, Jihwan Kwak, Subin Lim, Hyeonsu Bang, Hyojun Kim, and Taesup Moon.
\newblock Ss-il: Separated softmax for incremental learning.
\newblock In \emph{ICCV}, pages 844--853, 2021.

\bibitem[Aljundi et~al.(2018)Aljundi, Babiloni, Elhoseiny, Rohrbach, and Tuytelaars]{aljundi2018memory_reg_2}
Rahaf Aljundi, Francesca Babiloni, Mohamed Elhoseiny, Marcus Rohrbach, and Tinne Tuytelaars.
\newblock Memory aware synapses: Learning what (not) to forget.
\newblock In \emph{ECCV}, pages 139--154, 2018.

\bibitem[Bao et~al.(2021)Bao, Dong, Piao, and Wei]{bao2021beit_beit}
Hangbo Bao, Li~Dong, Songhao Piao, and Furu Wei.
\newblock Beit: Bert pre-training of image transformers.
\newblock \emph{arXiv preprint arXiv:2106.08254}, 2021.

\bibitem[Belouadah and Popescu(2019)]{belouadah2019il2m_cil2}
Eden Belouadah and Adrian Popescu.
\newblock Il2m: Class incremental learning with dual memory.
\newblock In \emph{ICCV}, pages 583--592, 2019.

\bibitem[Boschini et~al.(2022)Boschini, Buzzega, Bonicelli, Porrello, and Calderara]{boschini2022continual_ccic_gsscil2}
Matteo Boschini, Pietro Buzzega, Lorenzo Bonicelli, Angelo Porrello, and Simone Calderara.
\newblock Continual semi-supervised learning through contrastive interpolation consistency.
\newblock \emph{Pattern Recognition Letters}, 162:\penalty0 9--14, 2022.

\bibitem[Caron et~al.(2020)Caron, Misra, Mairal, Goyal, Bojanowski, and Joulin]{caron2020unsupervised_smav}
Mathilde Caron, Ishan Misra, Julien Mairal, Priya Goyal, Piotr Bojanowski, and Armand Joulin.
\newblock Unsupervised learning of visual features by contrasting cluster assignments.
\newblock \emph{NeurIPS}, 33:\penalty0 9912--9924, 2020.

\bibitem[Caron et~al.(2021)Caron, Touvron, Misra, J{\'e}gou, Mairal, Bojanowski, and Joulin]{caron2021emerging_dino}
Mathilde Caron, Hugo Touvron, Ishan Misra, Herv{\'e} J{\'e}gou, Julien Mairal, Piotr Bojanowski, and Armand Joulin.
\newblock Emerging properties in self-supervised vision transformers.
\newblock In \emph{ICCV}, pages 9650--9660, 2021.

\bibitem[Castro et~al.(2018)Castro, Mar{\'\i}n-Jim{\'e}nez, Guil, Schmid, and Alahari]{castro2018end_cil1}
Francisco~M Castro, Manuel~J Mar{\'\i}n-Jim{\'e}nez, Nicol{\'a}s Guil, Cordelia Schmid, and Karteek Alahari.
\newblock End-to-end incremental learning.
\newblock In \emph{ECCV}, pages 233--248, 2018.

\bibitem[Chamikara et~al.(2018)Chamikara, Bert{\'o}k, Liu, Camtepe, and Khalil]{chamikara2018efficient_privacy}
Mahawaga Arachchige~Pathum Chamikara, Peter Bert{\'o}k, Dongxi Liu, Seyit Camtepe, and Ibrahim Khalil.
\newblock Efficient data perturbation for privacy preserving and accurate data stream mining.
\newblock \emph{Pervasive and Mobile Computing}, 48:\penalty0 1--19, 2018.

\bibitem[Chaudhry et~al.(2018{\natexlab{a}})Chaudhry, Dokania, Ajanthan, and Torr]{chaudhry2018riemannian_reimAN_walk_data_centric3}
Arslan Chaudhry, Puneet~K Dokania, Thalaiyasingam Ajanthan, and Philip~HS Torr.
\newblock Riemannian walk for incremental learning: Understanding forgetting and intransigence.
\newblock In \emph{ECCV}, pages 532--547, 2018{\natexlab{a}}.

\bibitem[Chaudhry et~al.(2018{\natexlab{b}})Chaudhry, Ranzato, Rohrbach, and Elhoseiny]{chaudhry2018efficient_agem_data_centric2}
Arslan Chaudhry, Marc'Aurelio Ranzato, Marcus Rohrbach, and Mohamed Elhoseiny.
\newblock Efficient lifelong learning with a-gem.
\newblock \emph{arXiv preprint arXiv:1812.00420}, 2018{\natexlab{b}}.

\bibitem[Chen et~al.(2022)Chen, Jiang, Wang, Wan, Wang, and Long]{chen2022debiased_ss1}
Baixu Chen, Junguang Jiang, Ximei Wang, Pengfei Wan, Jianmin Wang, and Mingsheng Long.
\newblock Debiased self-training for semi-supervised learning.
\newblock \emph{NeurIPS}, 35:\penalty0 32424--32437, 2022.

\bibitem[Chen et~al.(2020)Chen, Kornblith, Norouzi, and Hinton]{chen2020simple_simclr}
Ting Chen, Simon Kornblith, Mohammad Norouzi, and Geoffrey Hinton.
\newblock A simple framework for contrastive learning of visual representations.
\newblock In \emph{ICML}, pages 1597--1607. PMLR, 2020.

\bibitem[Chen et~al.(2024)Chen, Ding, Wang, Xin, Mo, Wang, Han, Luo, Zeng, and Wang]{chen2024context_cae}
Xiaokang Chen, Mingyu Ding, Xiaodi Wang, Ying Xin, Shentong Mo, Yunhao Wang, Shumin Han, Ping Luo, Gang Zeng, and Jingdong Wang.
\newblock Context autoencoder for self-supervised representation learning.
\newblock \emph{International Journal of Computer Vision}, 132\penalty0 (1):\penalty0 208--223, 2024.

\bibitem[Chen and He(2021)]{chen2021exploring_simsiam}
Xinlei Chen and Kaiming He.
\newblock Exploring simple siamese representation learning.
\newblock In \emph{CVPR}, pages 15750--15758, 2021.

\bibitem[Deng et~al.(2009)Deng, Dong, Socher, Li, Li, and Fei-Fei]{imagenet_cvpr09}
J.~Deng, W.~Dong, R.~Socher, L.-J. Li, K.~Li, and L.~Fei-Fei.
\newblock {ImageNet: A Large-Scale Hierarchical Image Database}.
\newblock In \emph{CVPR}, 2009.

\bibitem[Dhar et~al.(2019)Dhar, Singh, Peng, Wu, and Chellappa]{dhar2019learning_algo_feat_2}
Prithviraj Dhar, Rajat~Vikram Singh, Kuan-Chuan Peng, Ziyan Wu, and Rama Chellappa.
\newblock Learning without memorizing.
\newblock In \emph{CVPR}, pages 5138--5146, 2019.

\bibitem[Dosovitskiy et~al.(2020)Dosovitskiy, Beyer, Kolesnikov, Weissenborn, Zhai, Unterthiner, Dehghani, Minderer, Heigold, Gelly, et~al.]{dosovitskiy2020image_vit}
Alexey Dosovitskiy, Lucas Beyer, Alexander Kolesnikov, Dirk Weissenborn, Xiaohua Zhai, Thomas Unterthiner, Mostafa Dehghani, Matthias Minderer, Georg Heigold, Sylvain Gelly, et~al.
\newblock An image is worth 16x16 words: Transformers for image recognition at scale.
\newblock \emph{arXiv preprint arXiv:2010.11929}, 2020.

\bibitem[Douillard et~al.(2020)Douillard, Cord, Ollion, Robert, and Valle]{douillard2020podnet_algo_feat_4}
Arthur Douillard, Matthieu Cord, Charles Ollion, Thomas Robert, and Eduardo Valle.
\newblock Podnet: Pooled outputs distillation for small-tasks incremental learning.
\newblock In \emph{ECCV}, pages 86--102. Springer, 2020.

\bibitem[Gao et~al.(2022)Gao, Zhao, Ghanem, and Zhang]{gao2022r_algo_relation_1}
Qiankun Gao, Chen Zhao, Bernard Ghanem, and Jian Zhang.
\newblock R-dfcil: Relation-guided representation learning for data-free class incremental learning.
\newblock In \emph{ECCV}, pages 423--439. Springer, 2022.

\bibitem[Gidaris et~al.(2018)Gidaris, Singh, and Komodakis]{gidaris2018unsupervised_rotations}
Spyros Gidaris, Praveer Singh, and Nikos Komodakis.
\newblock Unsupervised representation learning by predicting image rotations.
\newblock \emph{arXiv preprint arXiv:1803.07728}, 2018.

\bibitem[Golab and {\"O}zsu(2003)]{golab2003issues_data_continous1}
Lukasz Golab and M~Tamer {\"O}zsu.
\newblock Issues in data stream management.
\newblock \emph{ACM Sigmod Record}, 32\penalty0 (2):\penalty0 5--14, 2003.

\bibitem[Gomes et~al.(2017)Gomes, Barddal, Enembreck, and Bifet]{gomes2017survey_data_continoius2}
Heitor~Murilo Gomes, Jean~Paul Barddal, Fabr{\'\i}cio Enembreck, and Albert Bifet.
\newblock A survey on ensemble learning for data stream classification.
\newblock \emph{ACM Computing Surveys (CSUR)}, 50\penalty0 (2):\penalty0 1--36, 2017.

\bibitem[Goodfellow et~al.(2013)Goodfellow, Mirza, Xiao, Courville, and Bengio]{goodfellow2013empirical}
Ian~J Goodfellow, Mehdi Mirza, Da~Xiao, Aaron Courville, and Yoshua Bengio.
\newblock An empirical investigation of catastrophic forgetting in gradient-based neural networks.
\newblock \emph{arXiv preprint arXiv:1312.6211}, 2013.

\bibitem[Grill et~al.(2020)Grill, Strub, Altch{\'e}, Tallec, Richemond, Buchatskaya, Doersch, Avila~Pires, Guo, Gheshlaghi~Azar, et~al.]{grill2020bootstrap_byol}
Jean-Bastien Grill, Florian Strub, Florent Altch{\'e}, Corentin Tallec, Pierre Richemond, Elena Buchatskaya, Carl Doersch, Bernardo Avila~Pires, Zhaohan Guo, Mohammad Gheshlaghi~Azar, et~al.
\newblock Bootstrap your own latent-a new approach to self-supervised learning.
\newblock \emph{NeurIPS}, 33:\penalty0 21271--21284, 2020.

\bibitem[He et~al.(2020)He, Fan, Wu, Xie, and Girshick]{he2020momentum_moco}
Kaiming He, Haoqi Fan, Yuxin Wu, Saining Xie, and Ross Girshick.
\newblock Momentum contrast for unsupervised visual representation learning.
\newblock In \emph{CVPR}, pages 9729--9738, 2020.

\bibitem[He et~al.(2022)He, Chen, Xie, Li, Doll{\'a}r, and Girshick]{he2022masked_mae}
Kaiming He, Xinlei Chen, Saining Xie, Yanghao Li, Piotr Doll{\'a}r, and Ross Girshick.
\newblock Masked autoencoders are scalable vision learners.
\newblock In \emph{CVPR}, pages 16000--16009, 2022.

\bibitem[Hinton et~al.(2015)Hinton, Vinyals, and Dean]{hinton2015distilling_kd_hinton}
Geoffrey Hinton, Oriol Vinyals, and Jeff Dean.
\newblock Distilling the knowledge in a neural network.
\newblock \emph{arXiv preprint arXiv:1503.02531}, 2015.

\bibitem[Hou et~al.(2018)Hou, Pan, Loy, Wang, and Lin]{hou2018lifelong_algo_logit_3}
Saihui Hou, Xinyu Pan, Chen~Change Loy, Zilei Wang, and Dahua Lin.
\newblock Lifelong learning via progressive distillation and retrospection.
\newblock In \emph{ECCV}, pages 437--452, 2018.

\bibitem[Hou et~al.(2019)Hou, Pan, Loy, Wang, and Lin]{hou2019learning_algo_feat_1}
Saihui Hou, Xinyu Pan, Chen~Change Loy, Zilei Wang, and Dahua Lin.
\newblock Learning a unified classifier incrementally via rebalancing.
\newblock In \emph{CVPR}, pages 831--839, 2019.

\bibitem[Isele and Cosgun(2018)]{isele2018selective_experience_data_centric4}
David Isele and Akansel Cosgun.
\newblock Selective experience replay for lifelong learning.
\newblock In \emph{AAAI}, volume~32, 2018.

\bibitem[Kalla and Biswas(2022)]{kalla2022s3c}
Jayateja Kalla and Soma Biswas.
\newblock S3c: Self-supervised stochastic classifiers for few-shot class-incremental learning.
\newblock In \emph{ECCV}, pages 432--448. Springer, 2022.

\bibitem[Kalla and Biswas(2024)]{kalla2024robust}
Jayateja Kalla and Soma Biswas.
\newblock Robust feature learning and global variance-driven classifier alignment for long-tail class incremental learning.
\newblock In \emph{WACV}, pages 32--41, 2024.

\bibitem[Kalla et~al.(2023)Kalla, Punia, Dutta, and Biswas]{kalla2023generalized}
Jayateja Kalla, Prishruit Punia, Titir Dutta, and Soma Biswas.
\newblock Generalized semi-supervised class incremental learning in presence of outliers.
\newblock \emph{Multimedia Tools and Applications}, pages 1--17, 2023.

\bibitem[Kang et~al.(2022{\natexlab{a}})Kang, Yoon, Madjid, Hwang, and Yoo]{kang2022soft_fscil4}
Haeyong Kang, Jaehong Yoon, Sultan Rizky~Hikmawan Madjid, Sung~Ju Hwang, and Chang~D Yoo.
\newblock On the soft-subnetwork for few-shot class incremental learning.
\newblock \emph{arXiv preprint arXiv:2209.07529}, 2022{\natexlab{a}}.

\bibitem[Kang et~al.(2022{\natexlab{b}})Kang, Park, and Han]{kang2022class_algo_feat_3}
Minsoo Kang, Jaeyoo Park, and Bohyung Han.
\newblock Class-incremental learning by knowledge distillation with adaptive feature consolidation.
\newblock In \emph{CVPR}, pages 16071--16080, 2022{\natexlab{b}}.

\bibitem[Kang et~al.(2023)Kang, Fini, Nabi, Ricci, and Alahari]{kang2023soft_nncil_gsscil3}
Zhiqi Kang, Enrico Fini, Moin Nabi, Elisa Ricci, and Karteek Alahari.
\newblock A soft nearest-neighbor framework for continual semi-supervised learning.
\newblock In \emph{ICCV}, pages 11868--11877, 2023.

\bibitem[Khare et~al.(2021)Khare, Cao, and Rehg]{khare2021unsupervised_cil}
Shivam Khare, Kun Cao, and James Rehg.
\newblock Unsupervised class-incremental learning through confusion.
\newblock \emph{arXiv preprint arXiv:2104.04450}, 2021.

\bibitem[Kirkpatrick et~al.(2017)Kirkpatrick, Pascanu, Rabinowitz, Veness, Desjardins, Rusu, Milan, Quan, Ramalho, Grabska-Barwinska, et~al.]{kirkpatrick2017overcoming_reg_3}
James Kirkpatrick, Razvan Pascanu, Neil Rabinowitz, Joel Veness, Guillaume Desjardins, Andrei~A Rusu, Kieran Milan, John Quan, Tiago Ramalho, Agnieszka Grabska-Barwinska, et~al.
\newblock Overcoming catastrophic forgetting in neural networks.
\newblock \emph{Proceedings of the national academy of sciences}, 114\penalty0 (13):\penalty0 3521--3526, 2017.

\bibitem[Krempl et~al.(2014)Krempl, {\v{Z}}liobaite, Brzezi{\'n}ski, H{\"u}llermeier, Last, Lemaire, Noack, Shaker, Sievi, Spiliopoulou, et~al.]{krempl2014open_storage}
Georg Krempl, Indre {\v{Z}}liobaite, Dariusz Brzezi{\'n}ski, Eyke H{\"u}llermeier, Mark Last, Vincent Lemaire, Tino Noack, Ammar Shaker, Sonja Sievi, Myra Spiliopoulou, et~al.
\newblock Open challenges for data stream mining research.
\newblock \emph{ACM SIGKDD explorations newsletter}, 16\penalty0 (1):\penalty0 1--10, 2014.

\bibitem[Krizhevsky et~al.(2009)Krizhevsky, Hinton, et~al.]{krizhevsky2009learning_cifar}
Alex Krizhevsky, Geoffrey Hinton, et~al.
\newblock Learning multiple layers of features from tiny images.
\newblock 2009.

\bibitem[Larsson et~al.(2016)Larsson, Maire, and Shakhnarovich]{larsson2016learning_color1}
Gustav Larsson, Michael Maire, and Gregory Shakhnarovich.
\newblock Learning representations for automatic colorization.
\newblock In \emph{ECCV}, pages 577--593. Springer, 2016.

\bibitem[Larsson et~al.(2017)Larsson, Maire, and Shakhnarovich]{larsson2017colorization_color2}
Gustav Larsson, Michael Maire, and Gregory Shakhnarovich.
\newblock Colorization as a proxy task for visual understanding.
\newblock In \emph{CVPR}, pages 6874--6883, 2017.

\bibitem[Lee et~al.(2020)Lee, Hwang, and Shin]{selfsup}
Hankook Lee, Sung~Ju Hwang, and Jinwoo Shin.
\newblock Self-supervised label augmentation via input transformations.
\newblock In \emph{ICML}, pages 5714--5724. PMLR, 2020.

\bibitem[Li and Hoiem(2017)]{li2017learning_algo_logit_2}
Zhizhong Li and Derek Hoiem.
\newblock Learning without forgetting.
\newblock \emph{IEEE TPAMI}, 40\penalty0 (12):\penalty0 2935--2947, 2017.

\bibitem[Liu et~al.(2022)Liu, Hu, Cao, Bagdanov, Li, and Cheng]{liu2022long}
Xialei Liu, Yu-Song Hu, Xu-Sheng Cao, Andrew~D Bagdanov, Ke~Li, and Ming-Ming Cheng.
\newblock Long-tailed class incremental learning.
\newblock In \emph{ECCV}, pages 495--512. Springer, 2022.

\bibitem[Lopez-Paz and Ranzato(2017)]{lopez2017gradient_gem_data_centric1}
David Lopez-Paz and Marc'Aurelio Ranzato.
\newblock Gradient episodic memory for continual learning.
\newblock \emph{NeurIPS}, 30, 2017.

\bibitem[Noroozi and Favaro(2016)]{noroozi2016unsupervised_jigsaw}
Mehdi Noroozi and Paolo Favaro.
\newblock Unsupervised learning of visual representations by solving jigsaw puzzles.
\newblock In \emph{ECCV}, pages 69--84. Springer, 2016.

\bibitem[Peng et~al.(2022)Peng, Zhao, Wang, Li, and Lovell]{peng2022few_fscil3}
Can Peng, Kun Zhao, Tianren Wang, Meng Li, and Brian~C Lovell.
\newblock Few-shot class-incremental learning from an open-set perspective.
\newblock In \emph{ECCV}, pages 382--397. Springer, 2022.

\bibitem[Petit et~al.(2023)Petit, Popescu, Schindler, Picard, and Delezoide]{petit2023fetril_exemplar_free_cil3}
Gr{\'e}goire Petit, Adrian Popescu, Hugo Schindler, David Picard, and Bertrand Delezoide.
\newblock Fetril: Feature translation for exemplar-free class-incremental learning.
\newblock In \emph{WACV}, pages 3911--3920, 2023.

\bibitem[Pseudo-Label(2013)]{pseudo2013simple_ss2}
Dong-Hyun~Lee Pseudo-Label.
\newblock The simple and efficient semi-supervised learning method for deep neural networks.
\newblock In \emph{ICML 2013 Workshop: Challenges in Representation Learning}, pages 1--6, 2013.

\bibitem[Qiu et~al.(2023)Qiu, Fu, Zhang, Lei, and Peng]{qiu2023semantic_fscil5}
Wenhao Qiu, Sichao Fu, Jingyi Zhang, Chengxiang Lei, and Qinmu Peng.
\newblock Semantic-visual guided transformer for few-shot class-incremental learning.
\newblock \emph{arXiv preprint arXiv:2303.15494}, 2023.

\bibitem[Rebuffi et~al.(2017)Rebuffi, Kolesnikov, Sperl, and Lampert]{rebuffi2017icarl_algo_logit_1}
Sylvestre-Alvise Rebuffi, Alexander Kolesnikov, Georg Sperl, and Christoph~H Lampert.
\newblock icarl: Incremental classifier and representation learning.
\newblock In \emph{CVPR}, pages 2001--2010, 2017.

\bibitem[Rusu et~al.(2016)Rusu, Rabinowitz, Desjardins, Soyer, Kirkpatrick, Kavukcuoglu, Pascanu, and Hadsell]{rusu2016progressive_model_centric_3}
Andrei~A Rusu, Neil~C Rabinowitz, Guillaume Desjardins, Hubert Soyer, James Kirkpatrick, Koray Kavukcuoglu, Razvan Pascanu, and Raia Hadsell.
\newblock Progressive neural networks.
\newblock \emph{arXiv preprint arXiv:1606.04671}, 2016.

\bibitem[Selvaraju et~al.(2017)Selvaraju, Cogswell, Das, Vedantam, Parikh, and Batra]{selvaraju2017grad_cam}
Ramprasaath~R Selvaraju, Michael Cogswell, Abhishek Das, Ramakrishna Vedantam, Devi Parikh, and Dhruv Batra.
\newblock Grad-cam: Visual explanations from deep networks via gradient-based localization.
\newblock In \emph{ICCV}, pages 618--626, 2017.

\bibitem[Smith et~al.(2023)Smith, Karlinsky, Gutta, Cascante-Bonilla, Kim, Arbelle, Panda, Feris, and Kira]{smith2023coda_model_centric_6}
James~Seale Smith, Leonid Karlinsky, Vyshnavi Gutta, Paola Cascante-Bonilla, Donghyun Kim, Assaf Arbelle, Rameswar Panda, Rogerio Feris, and Zsolt Kira.
\newblock Coda-prompt: Continual decomposed attention-based prompting for rehearsal-free continual learning.
\newblock In \emph{CVPR}, pages 11909--11919, 2023.

\bibitem[Sohn et~al.(2020)Sohn, Berthelot, Carlini, Zhang, Zhang, Raffel, Cubuk, Kurakin, and Li]{sohn2020fixmatch_ss3}
Kihyuk Sohn, David Berthelot, Nicholas Carlini, Zizhao Zhang, Han Zhang, Colin~A Raffel, Ekin~Dogus Cubuk, Alexey Kurakin, and Chun-Liang Li.
\newblock Fixmatch: Simplifying semi-supervised learning with consistency and confidence.
\newblock \emph{NeurIPS}, 33:\penalty0 596--608, 2020.

\bibitem[Tao et~al.(2020)Tao, Hong, Chang, Dong, Wei, and Gong]{tao2020few_topic_fscil2}
Xiaoyu Tao, Xiaopeng Hong, Xinyuan Chang, Songlin Dong, Xing Wei, and Yihong Gong.
\newblock Few-shot class-incremental learning.
\newblock In \emph{CVPR}, pages 12183--12192, 2020.

\bibitem[Tarvainen and Valpola(2017)]{tarvainen2017mean_ss4}
Antti Tarvainen and Harri Valpola.
\newblock Mean teachers are better role models: Weight-averaged consistency targets improve semi-supervised deep learning results.
\newblock \emph{NeurIPS}, 30, 2017.

\bibitem[Wang et~al.(2022{\natexlab{a}})Wang, Zhou, Ye, and Zhan]{wang2022foster}
Fu-Yun Wang, Da-Wei Zhou, Han-Jia Ye, and De-Chuan Zhan.
\newblock Foster: Feature boosting and compression for class-incremental learning.
\newblock In \emph{ECCV}, pages 398--414. Springer, 2022{\natexlab{a}}.

\bibitem[Wang et~al.(2021)Wang, Yang, Li, Hong, Li, and Zhu]{wang2021ordisco_gsscil1}
Liyuan Wang, Kuo Yang, Chongxuan Li, Lanqing Hong, Zhenguo Li, and Jun Zhu.
\newblock Ordisco: Effective and efficient usage of incremental unlabeled data for semi-supervised continual learning.
\newblock In \emph{CVPR}, pages 5383--5392, 2021.

\bibitem[Wang et~al.(2022{\natexlab{b}})Wang, Zhang, Ebrahimi, Sun, Zhang, Lee, Ren, Su, Perot, Dy, et~al.]{wang2022dualprompt_model_centric_5}
Zifeng Wang, Zizhao Zhang, Sayna Ebrahimi, Ruoxi Sun, Han Zhang, Chen-Yu Lee, Xiaoqi Ren, Guolong Su, Vincent Perot, Jennifer Dy, et~al.
\newblock Dualprompt: Complementary prompting for rehearsal-free continual learning.
\newblock In \emph{ECCV}, pages 631--648. Springer, 2022{\natexlab{b}}.

\bibitem[Wang et~al.(2022{\natexlab{c}})Wang, Zhang, Lee, Zhang, Sun, Ren, Su, Perot, Dy, and Pfister]{wang2022learning_model_centric_4}
Zifeng Wang, Zizhao Zhang, Chen-Yu Lee, Han Zhang, Ruoxi Sun, Xiaoqi Ren, Guolong Su, Vincent Perot, Jennifer Dy, and Tomas Pfister.
\newblock Learning to prompt for continual learning.
\newblock In \emph{CVPR}, pages 139--149, 2022{\natexlab{c}}.

\bibitem[Xie et~al.(2022)Xie, Zhang, Cao, Lin, Bao, Yao, Dai, and Hu]{xie2022simmim}
Zhenda Xie, Zheng Zhang, Yue Cao, Yutong Lin, Jianmin Bao, Zhuliang Yao, Qi~Dai, and Han Hu.
\newblock Simmim: A simple framework for masked image modeling.
\newblock In \emph{CVPR}, pages 9653--9663, 2022.

\bibitem[Yan et~al.(2021)Yan, Xie, and He]{yan2021dynamically_model_centric2}
Shipeng Yan, Jiangwei Xie, and Xuming He.
\newblock Der: Dynamically expandable representation for class incremental learning.
\newblock In \emph{CVPR}, pages 3014--3023, 2021.

\bibitem[Yoon et~al.(2017)Yoon, Yang, Lee, and Hwang]{yoon2017lifelong_model_centric1}
Jaehong Yoon, Eunho Yang, Jeongtae Lee, and Sung~Ju Hwang.
\newblock Lifelong learning with dynamically expandable networks.
\newblock \emph{arXiv preprint arXiv:1708.01547}, 2017.

\bibitem[Yu et~al.(2020)Yu, Twardowski, Liu, Herranz, Wang, Cheng, Jui, and Weijer]{yu2020semantic_algo_relation_2}
Lu~Yu, Bartlomiej Twardowski, Xialei Liu, Luis Herranz, Kai Wang, Yongmei Cheng, Shangling Jui, and Joost van~de Weijer.
\newblock Semantic drift compensation for class-incremental learning.
\newblock In \emph{CVPR}, pages 6982--6991, 2020.

\bibitem[Zenke et~al.(2017)Zenke, Poole, and Ganguli]{zenke2017continual_SI_reg_1}
Friedemann Zenke, Ben Poole, and Surya Ganguli.
\newblock Continual learning through synaptic intelligence.
\newblock In \emph{ICML}, pages 3987--3995. PMLR, 2017.

\bibitem[Zhou et~al.(2023)Zhou, Wang, Qi, Ye, Zhan, and Liu]{zhou2023deep_deep_cil_survey}
Da-Wei Zhou, Qi-Wei Wang, Zhi-Hong Qi, Han-Jia Ye, De-Chuan Zhan, and Ziwei Liu.
\newblock Deep class-incremental learning: A survey.
\newblock \emph{arXiv preprint arXiv:2302.03648}, 2023.

\bibitem[Zhu et~al.(2021)Zhu, Zhang, Wang, Yin, and Liu]{zhu2021prototype_pass_exemplar_free_cil1}
Fei Zhu, Xu-Yao Zhang, Chuang Wang, Fei Yin, and Cheng-Lin Liu.
\newblock Prototype augmentation and self-supervision for incremental learning.
\newblock In \emph{CVPR}, pages 5871--5880, 2021.

\bibitem[Zhu et~al.(2022)Zhu, Zhai, Cao, Luo, and Zha]{zhu2022self_exemplar_free_cil2}
Kai Zhu, Wei Zhai, Yang Cao, Jiebo Luo, and Zheng-Jun Zha.
\newblock Self-sustaining representation expansion for non-exemplar class-incremental learning.
\newblock In \emph{CVPR}, pages 9296--9305, 2022.

\end{thebibliography}
\end{document}